\documentclass[10pt,twocolumn,letterpaper]{article}

\usepackage{cvpr}
\usepackage{times}
\usepackage{epsfig}
\usepackage{graphicx}
\usepackage{amsmath}
\usepackage{amssymb}

\usepackage{multirow}

\usepackage{pdfrender}

\usepackage{enumerate}
\usepackage{bm}
\usepackage{mathtools} 

\usepackage{subfig}
\usepackage{array}

\newcolumntype{x}[1]{>{\centering\arraybackslash}p{#1pt}}
\newcommand{\tablestyle}[2]{\setlength{\tabcolsep}{#1}\renewcommand{\arraystretch}{#2}\centering\footnotesize}
\usepackage[font=small]{caption}

\usepackage{algorithm}
\usepackage{algpseudocode}

\newcommand*\inlineimage[1]{\raisebox{-0.14\baselineskip}{\includegraphics[height=0.95\baselineskip]{#1}}}
\newcommand{\tieimg}{\inlineimage{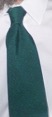}}

\usepackage[pagebackref=true,breaklinks=true,letterpaper=true,colorlinks,bookmarks=false]{hyperref}

\cvprfinalcopy 

\def\cvprPaperID{4748} 

\ifcvprfinal\fi
\begin{document}


\title{Counterfactual Samples Synthesizing for Robust Visual Question Answering}

\author{Long Chen$^1$\thanks{Long Chen and Xin Yan are co-first authors with equal contributions.} \quad Xin Yan$^1$\footnotemark[1] \quad Jun Xiao$^1$\thanks{Corresponding author.} \quad Hanwang Zhang$^2$ \quad Shiliang Pu$^3$ \quad Yueting Zhuang$^1$ \\
$^1$DCD Lab, College of Computer Science, Zhejiang University \\ $^2$MReaL Lab, Nanyang Technological University \qquad $^3$Hikvision Research Institute \\
}

\maketitle
\thispagestyle{empty}

\begin{abstract}
	Despite Visual Question Answering (VQA) has realized impressive progress over the last few years, today's VQA models tend to capture superficial linguistic correlations in the train set and fail to generalize to the test set with different QA distributions. To reduce the language biases, several recent works introduce an auxiliary question-only model to regularize the training of targeted VQA model, and achieve dominating performance on VQA-CP. However, since the complexity of design, current methods are unable to equip the ensemble-based models with two indispensable characteristics of an ideal VQA model: 1) visual-explainable: the model should rely on the right visual regions when making decisions. 2) question-sensitive: the model should be sensitive to the linguistic variations in question. To this end, we propose a model-agnostic Counterfactual Samples Synthesizing (CSS) training scheme. The CSS generates numerous counterfactual training samples by masking critical objects in images or words in questions, and assigning different ground-truth answers. After training with the complementary samples (\ie, the original and generated samples), the VQA models are forced to focus on all critical objects and words, which significantly improves both visual-explainable and question-sensitive abilities. In return, the performance of these models is further boosted. Extensive ablations have shown the effectiveness of CSS. Particularly, by building on top of the model LMH~\cite{clark2019don}, we achieve a record-breaking performance of 58.95\% on VQA-CP v2, with 6.5\% gains.\footnote{Codes: \href{https://github.com/yanxinzju/CSS-VQA}{https://github.com/yanxinzju/CSS-VQA}}
\end{abstract}
\vspace{-0.5em}

\begin{figure}[tbp]
	\centering
	\includegraphics[width=0.95\linewidth]{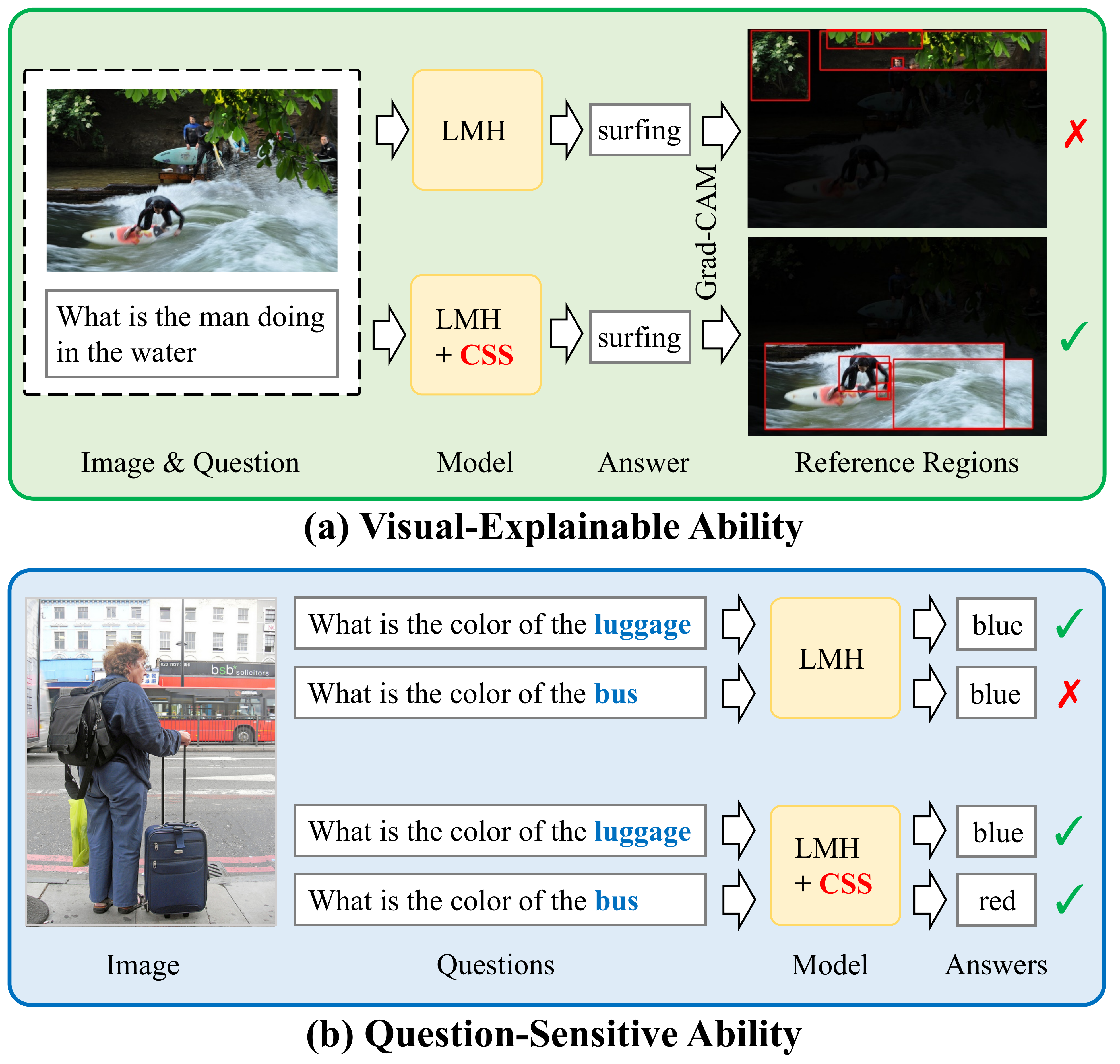}
	\caption{The two indispensable characteristics of an ideal VQA model. (a) \textbf{visual-explainable ability}: the model not only needs to predict correct answer (\eg, ``surfing"), but also relies on the right reference regions when making this prediction. (b) \textbf{question-sensitive ability}: the model should be sensitive to the linguistic variations, \eg, after replacing the critical word ``luggage" with ``bus", the predicted answers of two questions should be different.}
	\label{fig:1}
\end{figure}

\section{Introduction} \label{sec:intro}
Visual Question Answering (VQA), \ie, answering natural language questions about the visual content, is one of the core techniques towards complete AI. With the release of multiple large scale VQA datasets (\eg, VQA v1~\cite{antol2015vqa} and v2~\cite{goyal2017making}), VQA has received unprecedented attention and hundreds of models have been developed. However, since the inevitable annotation artifacts in the real image datasets, today's VQA models always over-rely on superficial linguistic correlations (\ie, language biases)~\cite{agrawal2016analyzing, zhang2016yin, johnson2017clevr, goyal2017making}. For example, a model answering ``2" for all ``\textit{how many X}" questions can still get satisfactory performance regardless of the \textit{X}. Recently, to disentangle the bias factors and clearly monitor the progress of VQA research, a diagnostic benchmark VQA-CP (VQA under Changing Priors)~\cite{agrawal2018don} has been proposed. The VQA-CP deliberately has different question-answer distributions in the train and test splits. The performance of many state-of-the-art VQA models~\cite{andreas2016neural, fukui2016multimodal, yang2016stacked, anderson2018bottom} drop significantly on VQA-CP compared to other datasets.

Currently, the prevailing solutions to mitigate the bias issues are \textbf{ensemble-based} methods: they introduce an auxiliary question-only model to regularize the training of targeted VQA model. Specifically, these methods can further be grouped into two sub-types: 1) \emph{adversary-based}~\cite{ramakrishnan2018overcoming, grand2019adversarial, belinkov2019don}: they train two models in an adversarial manner~\cite{goodfellow2014generative, chen2018zero}, \ie, minimizing the loss of VQA model while maximizing the loss of question-only model. Since the two models are designed to share the same question encoder, the adversary-based methods aim to reduce the language biases by learning a bias-neutral question representation. Unfortunately, the adversarial training scheme brings significant noise into gradients and results in an unstable training process~\cite{grand2019adversarial}. 2) \emph{fusion-based}~\cite{cadene2019rubi, clark2019don, mahabadi2019simple}: they late fuse the predicted answer distributions of the two models, and derive the training gradients based on the fused answer distributions. The design philosophy of the fusion-based methods, is to let the targeted VQA model focuses more on the samples, which cannot be answered correctly by the question-only model.

\begin{figure}[tbp]
	\centering
	\includegraphics[width=0.95\linewidth]{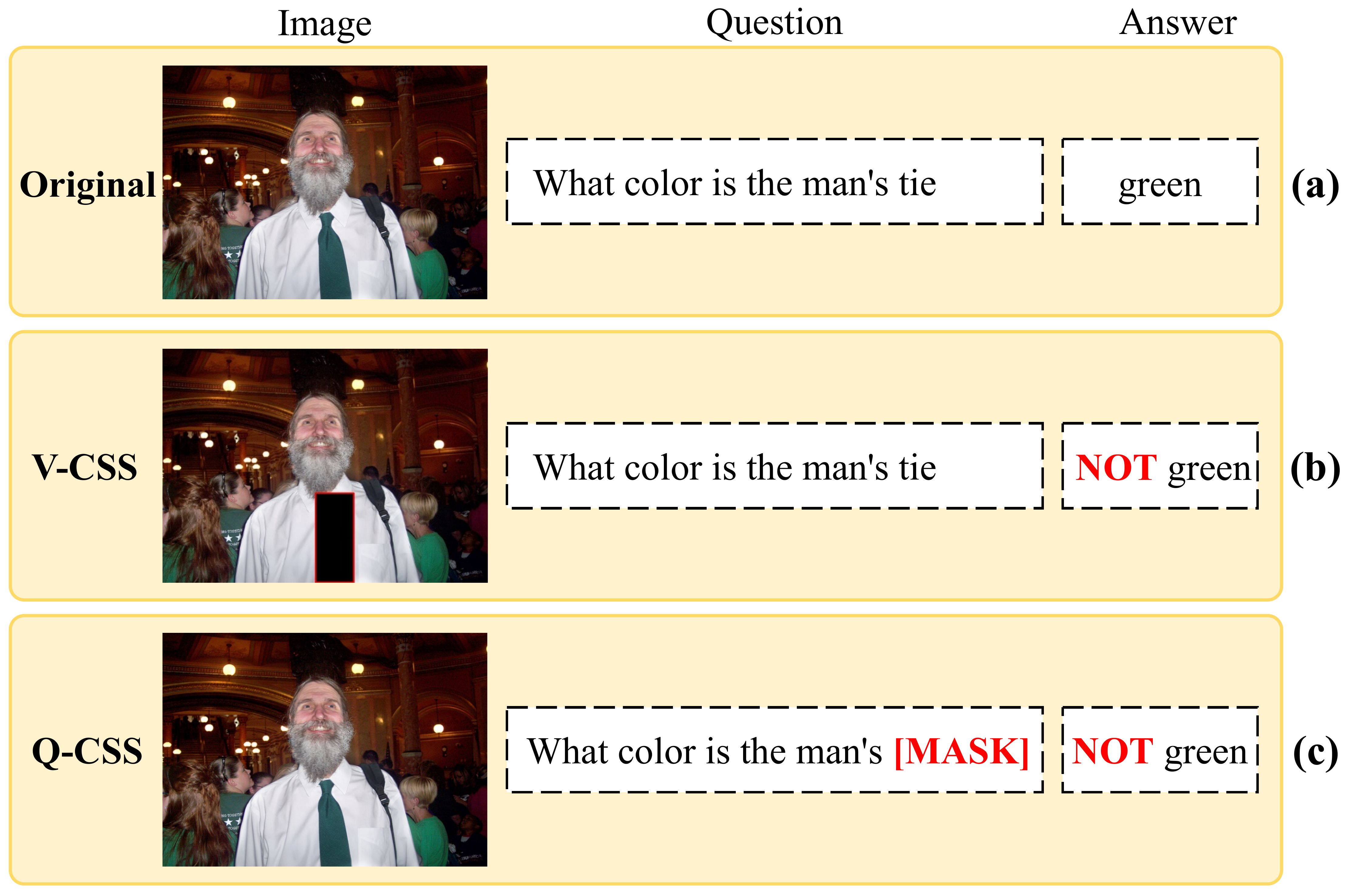}
	\vspace{-0.5em}
	\caption{(a): A training sample from the VQA-CP. (b): The synthesized training sample by V-CSS. It masks ciritcal objects (\eg, ``tie") in image and assigns different ground-truth answers (``not green"). (c): The synthesized training sample by Q-CSS. It replaces critical words (\eg, ``tie") with special token ``[MASK]" in question and assigns different ground-truth answers (``not green").}
	\label{fig:2}
\end{figure}

Although the ensemble-based methods have dominated the performance on VQA-CP, it is worth noting that current methods fail to equip them with two indispensable characteristics of an ideal VQA model: 1) \textbf{visual-explainable}: the model should rely on the right visual regions when making decisions, \ie, right for the right reasons~\cite{ross2017right}. As shown in Figure~\ref{fig:1} (a), although both two models can predict the correct answer ``surfing", they actually refer to totally different reference regions when making this answer prediction. 2) \textbf{question-sensitive}: the model should be sensitive to the linguistic variations in question. As shown in Figure~\ref{fig:1} (b), for two questions with similar sentence structure (\eg,  only replacing word ``luggage" with ``bus"), if the meanings of two questions are different, the model should perceive the discrepancy and make corresponding predictions.

In this paper, we propose a novel model-agnostic Counterfactual Samples Synthesizing (CSS) training scheme. The CSS serves as a plug-and-play component to improve the VQA models' visual-explainable and question-sensitive abilities, even for complex ensemble-based methods. As shown in Figure~\ref{fig:2}, CSS consists of two different types of samples synthesizing mechanisms: V-CSS and Q-CSS. For V-CSS, it synthesizes a counterfactual image by masking critical objects in the original image. By ``critical", we mean that these objects are important in answering a certain question (\eg, object~~\tieimg~~for the question ``\texttt{what color is the man's tie}"). Then, the counterfactual image and original question compose a new image-question (VQ) pair. For Q-CSS, it synthesizes a counterfactual question by replacing critical words in the original question with a special token ``[MASK]". Similarly, the counterfactual querstion and original image compose a new VQ pair. Given a VQ pair (from V-CSS or Q-CSS), a standard VQA training sample triplet still needs the corresponding ground-truth answers. To avoid the expensive manual annotations, we design a dynamic answer assigning mechanism to approximate ground-truth answers for all synthesized VQ pairs (\eg, ``\texttt{not green}" in Figure~\ref{fig:2}). Then, we train the VQA models with all original and synthesized samples. After training with numerous complementary samples, the VQA models are forced to focus on critical objects and words.

Extensive ablations including both qualitative and quantitative results have demonstrated the effectiveness of CSS. The CSS can be seamlessly incorporated into the ensemble-based methods, which not only improves their both visual-explainable and question-sensitive abilities, but also consistently boosts the performance on VQA-CP. Particularly, by building of top on model LMH~\cite{clark2019don}, we achieve a new record-breaking performance of 58.95\% on VQA-CP v2.

\section{Related Work}

\textbf{Language Biases in VQA.}
Despite VQA is a multimodal task, a large body of research~\cite{jabri2016revisiting, agrawal2016analyzing, zhang2016yin, goyal2017making} has shown the existence of language biases in VQA. There are two main solutions to reduce the language biases:

\noindent\textit{1. Balancing Datasets to Reduce Biases.} The most straightforward solution is to create more balanced datasets. For example, Zhang~\etal~\cite{zhang2016yin} collected complementary abstract scenes with opposite answers for all binary questions. And Goyal~\etal~\cite{goyal2017making} extended this idea into real images and all types of questions. Although these ``balanced" datasets have reduced biases to some extent, the statistical biases from questions still can be leveraged~\cite{agrawal2018don}. As shown in the benchmark VQA-CP, the performance of numerous models drop significantly compared to these ``balanced" datasets. In this paper, we follow the same spirit of dataset balancing and train VQA models with more complementary samples. Especially, CSS doesn't need any extra manual annotations.

\noindent\textit{2. Designing Models to Reduce Biases.} Another solution is to design specific debiasing models. So far, the most effective debiasing models for VQA are ensemble-based methods~\cite{ramakrishnan2018overcoming, grand2019adversarial, belinkov2019don, cadene2019rubi, clark2019don, mahabadi2019simple}. In this paper, we propose a novel CSS training scheme, which can be seamlessly incorporated into the ensemble-based models to further reduce the biases.

\textbf{Visual-Explainable Ability in VQA Models.}
To improve visual-explainable ability, early works~\cite{qiao2018exploring, liu2017attention, zhang2019interpretable} directly apply human attention as supervision to guide the models' attention maps. However, since the existence of strong biases, even with appropriate attention maps, the remaining layers of network may still disregard the visual signal~\cite{selvaraju2019taking}. Thus, some recent works~\cite{selvaraju2019taking, wu2019self} utilize Grad-CAM~\cite{selvaraju2017grad} to obtain private contribution of each object to correct answers, and encourage the rank of all object contributions to be consistent with human annotations. Unfortunately, these models have two drawbacks: 1) They need extra human annotations. 2) The training is not end-to-end.

\textbf{Question-Sensitive Ability in VQA Models.}
If VQA systems really ``understand" the question, they should be sensitive to the linguistic variations in question. Surprisingly, to the best of our knowledge, there is only one work~\cite{shah2019cycle} has studied the influence of linguistic variations in VQA. Specifically, it designs a cycle-consistent loss between two dual tasks, and utilizes sampled noises to generate diverse questions. However, Shah~\etal~\cite{shah2019cycle} only considers the robustness to different rephrasings of questions. In contrast, we also encourage the model to perceive the difference of questions when changing some critical words.

\textbf{Counterfactual Training Samples for VQA.}
Some concurrent works~\cite{agarwal2019towards,pan2019question} also try to synthesize counterfactual samples for VQA. Different from these works that all resort to GAN~\cite{goodfellow2014generative} to generate images, CSS only mask critical objects or words, which is easier and more adoptable.

\section{Approach}

We consider the common formulation of VQA task as a multi-class classification problem. Without loss of generality, given a dataset $\mathcal{D} = \{I_i, Q_i, a_i \}^N_i$ consisting of triplets of images $I_i \in \mathcal{I}$, questions $Q_i \in \mathcal{Q}$ and answers $a_i \in \mathcal{A}$, VQA task learns a mapping $f_{vqa}: \mathcal{I} \times \mathcal{Q} \rightarrow [0, 1]^{|\mathcal{A}|}$, which produces an answer distribution given image-question pair. For simplicity, we omit subscript $i$ in the following sections.

In this section, we first introduce the base bottom-up top-down model~\cite{anderson2018bottom}, and the ensemble-based methods for debiasing in Section~\ref{sec:3.1}. Then, we introduce the details of the Counterfactual Samples Synthesizing (CSS) in Section~\ref{sec:3.2}.

\subsection{Preliminaries} \label{sec:3.1}

\noindent\textbf{Bottom-Up Top-Down (UpDn) Model.}
For each image $I$, the UpDn uses an image encoder $e_v$ to output a set of object features: $\bm{V} = \{\bm{v}_1, ..., \bm{v}_{n_v}\}$, where $\bm{v}_i$ is $i$-th object feature. For each question $Q$, the UpDn uses a question encoder $e_q$ to output a set of word features: $\bm{Q} = \{\bm{w}_1, ..., \bm{w}_{n_q}\}$, where $\bm{w}_j$ is $j$-th word feature. Then both $\bm{V}$ and $\bm{Q}$ are fed into the model $f_{vqa}$ to predict answer distributions:
\begin{equation} \label{eq:p_vqa}
	P_{vqa}(\bm{a}|I, Q) = f_{vqa}(\bm{V}, \bm{Q}).
\end{equation}
Model $f_{vqa}$ typically contains an attention mechanism~\cite{chen2017sca,niu2019recursive,ye2017video}, and it is trained with cross-entropy loss~\cite{tang2019learning,chen2019counterfactual}.

\begin{algorithm}[t]
	\caption{Ensemble-based Model (fusion-based)}\label{alg:VQA}
	\begin{algorithmic}[1]
		\Function {$\mathcal{\textcolor{blue}{VQA}}$}{$I, Q, a, cond$}
		\State $ \bm{V} \leftarrow e_v(I) $
		\State $ \bm{Q} \leftarrow e_q(Q) $
		\State $ P_{vqa}(\bm{a}) \leftarrow f_{vqa}(\bm{V}, \bm{Q}) $
		\State $ P_{q}(\bm{a}) \leftarrow f_{q}(\bm{Q})$      \Comment{question-only model}
		\State $ \hat{P}_{vqa}(\bm{a}) \leftarrow M(P_{vqa}(\bm{a}), P_{q}(\bm{a}))  $
		\State $ Loss \leftarrow \text{XE}(\hat{P}_{vqa}(\bm{a}), a)$ \Comment{update parameters}
		\If{$cond$}
		\State \textbf{return} $\bm{V}, \bm{Q}, P_{vqa}(\bm{a})$
		\EndIf 
		\EndFunction
	\end{algorithmic}
\end{algorithm}

\noindent\textbf{Ensemble-Based Models.} As we discussed in Section~\ref{sec:intro}, the ensemble-based models can be grouped into two sub-types: adversary-based and fusion-based. Since adversary-based models~\cite{ramakrishnan2018overcoming, grand2019adversarial, belinkov2019don} suffer severe unstable training and relatively worse performance, in this section, we only introduce the fusion-based models~\cite{cadene2019rubi, clark2019don, mahabadi2019simple}. As shown in Algorithm~\ref{alg:VQA}, they introduce an auxiliary question-only model $f_q$ which takes $\bm{Q}$ as input and predicts answer distribution:
\begin{equation}
P_{q}(\bm{a}|Q) = f_{q}(\bm{Q}).
\end{equation}

Then, they combine the two answer distributions and obtain a new answer distribution $\hat{P}_{vqa}(\bm{a})$ by a function $M$:
\begin{equation}
\hat{P}_{vqa}(\bm{a}|I, Q) = M(P_{vqa}(\bm{a}|I, Q), P_{q}(\bm{a}|Q)).
\end{equation}
In the training stage, the XE loss is computed based on the fused answer distribution $\hat{P}_{vqa}(\bm{a})$ and the training gradients are backpropagated through both $f_{vqa}$ and $f_q$. In test stage, only model $f_{vqa}$ is used as the plain VQA models.

\subsection{Counterfactual Samples Synthesizing (CSS)} \label{sec:3.2}

\begin{algorithm}[tbp]
	\caption{Counterfactual Samples Synthesizing}\label{alg:CSS}
	\begin{algorithmic}[1]
		\Function {$\mathcal{CSS}$}{$I, Q, a$}
		\State $ \bm{V}, \bm{Q}, P_{vqa}(\bm{a}) \leftarrow \mathcal{\textcolor{blue}{VQA}}(I, Q, a, \text{True})$
		\State $ cond \sim U[0, 1]$
		\If {$cond \geq \delta $}  \Comment{execute V-CSS}
			\State $ \mathcal{I} \leftarrow  \textsc{IO\_Sel}(I, Q) $
			\State $ s(a, \bm{v}_i) \leftarrow \mathcal{S}(P_{vqa}(a), \bm{v}_i)$
			\State $ I^+, I^- \leftarrow \textsc{CO\_Sel}(\mathcal{I}, \{s(a, \bm{v}_i) \}) $
			\State $ a^- \leftarrow \textsc{DA\_Ass}(I^+, Q, \mathcal{VQA}, a) $
			\State $ \mathcal{\textcolor{blue}{VQA}}(I^-, Q, a^-, \text{False})$
		\Else \Comment{execute Q-CSS}
			\State $ s(a, \bm{w}_i) \leftarrow \mathcal{S}(P_{vqa}(a), \bm{w}_i) $
			\State $ Q^+, Q^- \leftarrow \textsc{CW\_Sel}(\{s(a, \bm{w}_i)\})$
			\State $ a^- \leftarrow \textsc{DA\_Ass}(I, Q^+, \mathcal{VQA}, a) $
			\State $ \mathcal{\textcolor{blue}{VQA}}(I, Q^-, a^-, \text{False})$
		\EndIf
		\EndFunction
	\end{algorithmic}
\end{algorithm}

The overall structure of CSS training scheme is shown in Algorithm~\ref{alg:CSS}. Specifically, for any $\mathcal{\textcolor{blue}{VQA}}$ model, given a training sample $(I, Q, a)$, CSS consists of three main steps: 
\begin{enumerate}
	\vspace{-0.6em}
	\itemsep -0.2em
	\item Training $\mathcal{\textcolor{blue}{VQA}}$ model with original sample $(I, Q, a)$;
	\item Synthesizing a counterfactual sample $(I^-, Q, a^-)$ by V-CSS or $(I, Q^-, a^-)$ by Q-CSS;
	\item Training $\mathcal{\textcolor{blue}{VQA}}$ model with the counterfactual sample.
	\vspace{-0.6em}
\end{enumerate}

In the following, we introduce the details of V-CSS and Q-CSS (\ie, the second step). As shown in Algorithm~\ref{alg:CSS}, for each training sample, we only use one certain synthesizing mechanism, and $\delta$ is the trade-off weight (See Figure~\ref{fig:4} (c) for more details about the influence of different $\delta$).

\subsubsection{V-CSS} \label{sec:v-css}

We sequentially introduce all steps of V-CSS following its execution path (line 5 to 8 in Algorithm~\ref{alg:CSS}), which consists of four main steps: initial objects selection (\textsc{IO\_Sel}), object local contributions calculation, critical objects selection (\textsc{CO\_Sel}), and dynamic answer assigning (\textsc{DA\_Ass}).

\textbf{1. Initial Objects Selection (\textsc{IO\_Sel}).} In general, for any specific QA pair $(Q, a)$, only a few objects in image $I$ are related. To narrow the scope of critical objects selection, we first construct a smaller object set $\mathcal{I}$, and assume all objects in $\mathcal{I}$ are possibly important in answering this question. Since we lack annotations about the critical objects for each sample, we followed~\cite{wu2019self} to extract the objects which are highly related with the QA. Specifically, we first assign POS tags to each word in the QA using the spaCy POS tagger~\cite{honnibal2017spacy} and extract nouns in QA. Then, we calculate the cosine similarity between the GloVe~\cite{pennington2014glove} embedding of object categories and the extracted nouns, the similarity scores between all objects in $I$ and the QA are denoted as $\mathcal{SIM}$. We select $|\mathcal{I}|$ objects with the highest $\mathcal{SIM}$ scores as $\mathcal{I}$. 

\textbf{2. Object Local Contributions Calculation.} After obtaining the object set $\mathcal{I}$, we start to calculate the local contribution of each object to the predicted probability of ground-truth answer. Following recent works~\cite{jain2019attention, selvaraju2019taking, wu2019self} which utilize the modified Grad-CAM~\cite{selvaraju2017grad} to derive the local contribution of each participant, we calculate the contribution of $i$-th object feature to the ground-truth answer $a$ as:
\begin{equation} \label{eq:object_gradcam}
s(a, \bm{v}_i) = \mathcal{S}(P_{vqa}(a), \bm{v}_i) \coloneqq (\nabla_{\bm{v}_i} P_{vqa}(a))^T\mathbf{1},
\end{equation}
where $P_{vqa}(a)$ is the predicted answer probability of ground truth answer a,  $\bm{v}_i$ is $i$-th object feature, and $\mathbf{1}$ is an all-ones vector. Obviously, if the score $s(a, \bm{v}_i)$ is higher, the contributions of object $\bm{v}_i$ to answer $a$ is larger.

\begin{algorithm}[tbp]
	\caption{Dynamic Answer Assigning}\label{alg:daass}
	\begin{algorithmic}[1]
		\Function {$\textsc{DA\_Ass}$}{$I^+, Q^+, \mathcal{VQA}, a$}
		\State	$\mathcal{VQA}$.eval() \Comment{\textcolor{red}{don't} update parameters}
		\State  $ \_, \_, P_{vqa}^+(\bm{a}) \leftarrow \mathcal{VQA}(I^+, Q^+, a, \text{True}) $
		\State $ a^+ \leftarrow \text{top-N}(\text{argsort}_{a_i \in \mathcal{A}}(P_{vqa}^+(a_i)))$
		\State $ a^- \coloneqq \{a_i | a_i \in a, a_i \notin a^+ \} $ \Comment{$a$ is gt answer set}
		\State \textbf{return} $a^-$
		\EndFunction
	\end{algorithmic}
\end{algorithm}

\begin{figure}[tbp]
	\centering
	\includegraphics[width=0.95\linewidth]{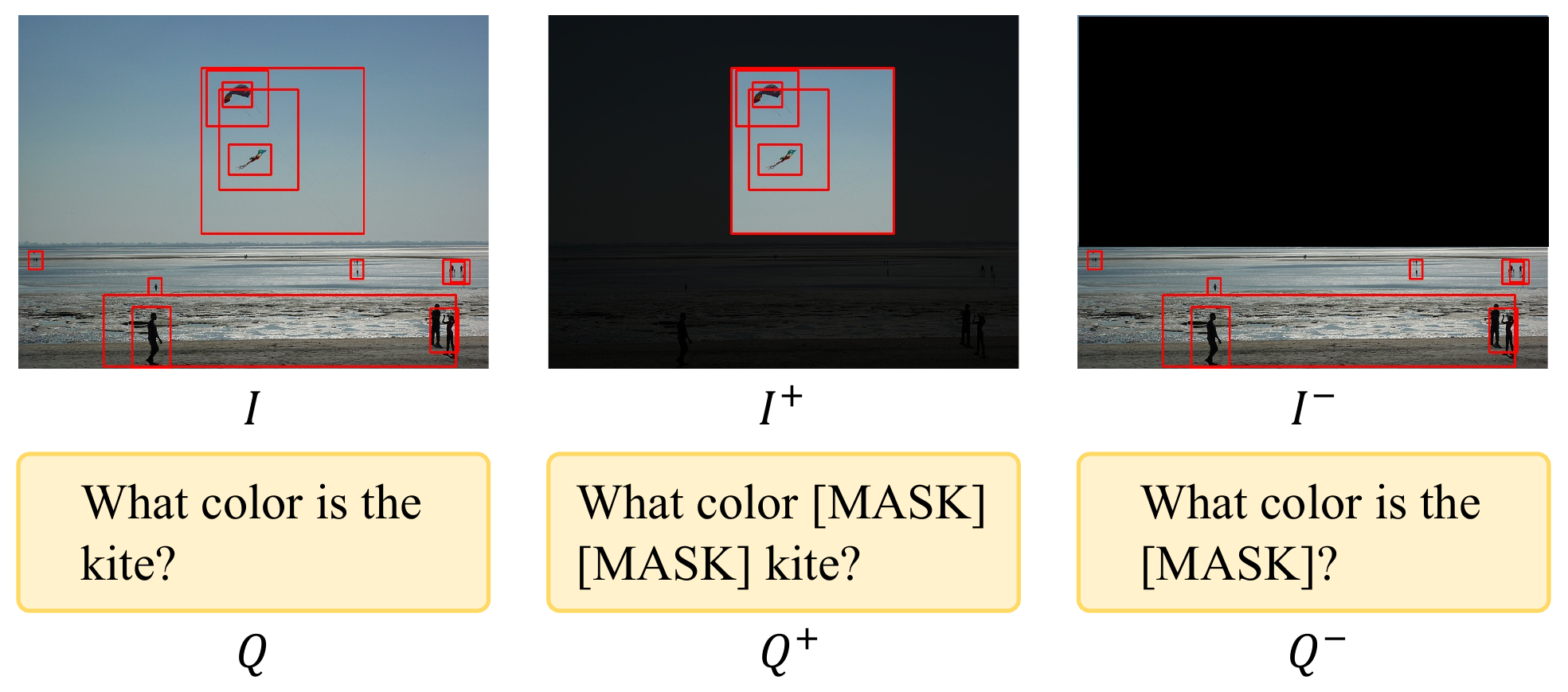}
	\vspace{-0.5em}
	\caption{An informal illustration example of the $I^+$, $I^-$, $Q^+$, and $Q^-$ in $\mathcal{CSS}$. For $I^+$ and $I^-$, they are two mutual exclusive object sets. For $Q^+$ and $Q^-$, we show the example when word "kite" is selected as critical word.
	}
	\label{fig:3}
\end{figure}

\textbf{3. Critical Objects Selection (\textsc{CO\_Sel}).} After obtaining the private contribution scores $s(a, \bm{v}_i)$ for all objects in $\mathcal{I}$, we select the top-K objects with highest scores as the critical object set $I^+$. The K is a dynamic number for each image, which is the smallest number meets Eq.~\eqref{eq:topk_objects}:
\begin{equation} \label{eq:topk_objects}
\sum_{\bm{v}_i \in I^+} \exp(s(a, \bm{v}_i)) / \sum_{\bm{v}_j \in \mathcal{I}} \exp(s(a, \bm{v}_j))  > \eta,
\end{equation}
where $\eta$ is a constant, we set $\eta=0.65$ in all experiments (See Figure~\ref{fig:4} for more details about the dynamic K setting).

Then, the counterfactual visual input $I^-$ is the absolute complement of set $I^+$ in set $I$, \ie, $I^- = I \backslash I^+$. We show an example of $I$, $I^+$, and $I^-$ in Figure~\ref{fig:3}.

\textbf{4. Dynamic Answer Assigning (\textsc{DA\_Ass}).} Given the counterfactual visual input $I^-$ and original question $Q$, we compose a new VQ pair ($I^-, Q$). To assign ground truth answers for VQ pair  ($I^-, Q$), we design a dynamic answer assigning (\textsc{DA\_Ass}) mechanism. The details of \textsc{DA\_Ass} are shown in Algorithm~\ref{alg:daass}. Specifically, we first feed another VQ pair ($I^+, Q$) into the $\mathcal{\textcolor{blue}{VQA}}$ model, and obtain the predicted answer distribution $P^+_{vqa}(\bm{a})$. Based on $P^+_{vqa}(\bm{a})$, we select the top-N answers with highest predicted probabilities as $a^+$. Then we define $ a^- \coloneqq \{a_i | a_i \in a, a_i \notin a^+ \} $. In an extreme case, if the model predicts all ground truth answer correctly for VQ pair ($I^+, Q$), \ie, $a \subset a^+$, then $a^-$ is a $\emptyset$, \ie, zero for all answer candidates. The basic motivation is that if current model can predict ground truth answer for ($I^+, Q$) (\ie, $I^+$ contains critial objects and $I^-$ not), the ground truth for ($I^-, Q$) should not contain original ground truth answers anymore, \eg, "\texttt{not green}" in Figure~\ref{fig:2}.

\subsubsection{Q-CSS}
All steps in Q-CSS are similar to V-CSS. Following its execution path (line 11 to 13 in Algorithm~\ref{alg:CSS}), it consists of word local contribution calculation, critical words selection (\textsc{CW\_Sel}), and dynamic answer assigning (\textsc{DA\_Ass}).

\textbf{1. Word Local Contribution Calculation.} Similar with the V-CSS (cf. Eq.~\eqref{eq:object_gradcam}), we calculate the contribution of $i$-th word feature to the ground-truth answer $a$ as:
\begin{equation} \label{eq:word_gradcam}
s(a, \bm{w}_i) = \mathcal{S}(P_{vqa}(a), \bm{w}_i) \coloneqq (\nabla_{\bm{w}_i} P_{vqa}(a))^T\mathbf{1}.
\end{equation}

\textbf{2. Critical Words Selection (\textsc{CW\_Sel}.)}  In this step, we first extract question-type words for each question $Q$\footnote{We use the default question-type annotations in VQA-CP dataset.} (\eg, "\texttt{what color}" in Figure~\ref{fig:3}). Then, we select top-K words with highest scores from the remaining sentence (except the question-type words) as critical words. The counterfactual question $Q^-$ is the sentence by replacing all critical words in $Q$ with a special token ``[MASK]". Meanwhile, the $Q^+$ is the sentence by replacing all other words (except question-type and critical words) with ``[MASK]". We show an example of $Q$, $Q^+$, and $Q^-$ in Figure~\ref{fig:3}.

\textbf{3. Dynamic Answer Assigning (\textsc{DA\_Ass}.)} This step is identical to the \textsc{DA\_Ass} in V-CSS, \ie, Algorithm~\ref{alg:daass}. For Q-CSS, the input for \textsc{DA\_Ass} is the VQ pair $(I, Q^+)$.

\section{Experiments}

\noindent\textbf{Settings.} We evaluated the proposed CSS for VQA mainly on the VQA-CP test set~\cite{agrawal2018don}. We also presented experimental results on the VQA v2 validation set~\cite{goyal2017making} for completeness. For model accuracies, we followed the standard  VQA evaluation metric~\cite{antol2015vqa}. For fair comparisons, we did all the same data preprocessing steps with the widely-used UpDn model~\cite{anderson2018bottom} using the publicly available reimplementation\footnote{https://github.com/hengyuan-hu/bottom-up-attention-vqa}.

\subsection{Ablative Studies }
\subsubsection{Hyperparameters of V-CSS and Q-CSS}
\begin{figure*}[htbp]
	\centering
	\includegraphics[width=0.95\linewidth]{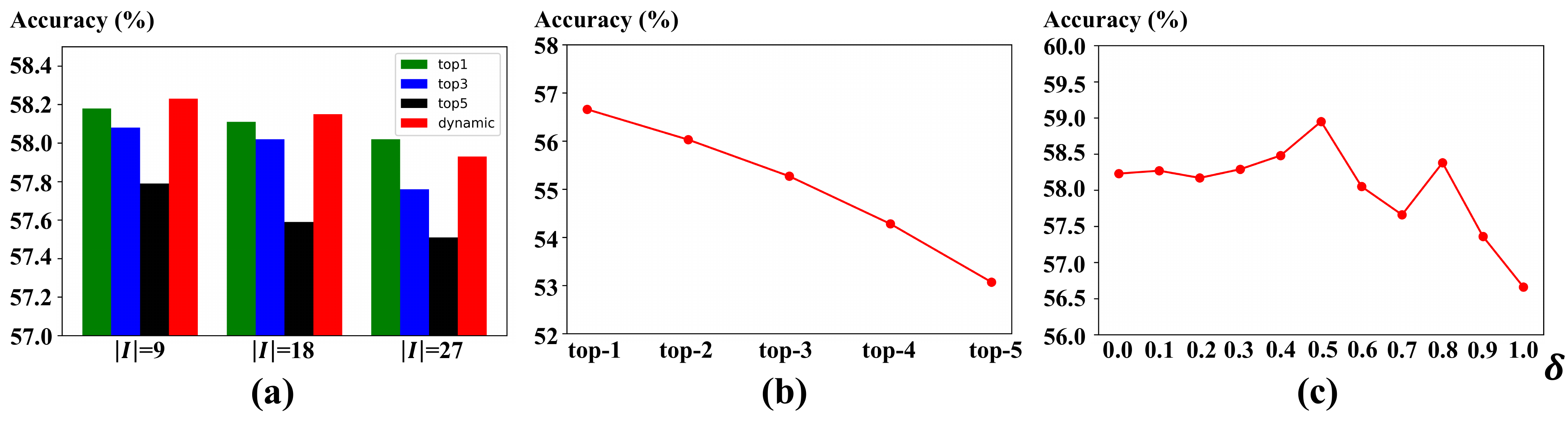}
	\vspace{-0.8em}
	\caption{\textbf{Ablations}. Accuracies (\%) on VQA-CP v2 test set of different hyperparameters settings of V-CSS or Q-CSS. (a) The results of different size of $\mathcal{I}$ and critical objects in V-CSS. All results come from model LMH+V-CSS. (b) The results of different size of critical words in Q-CSS. All results come from model LMH+Q-CSS. (c) The results of different $\delta$. All results come from model LMH+V-CSS+Q-CSS. }
	\label{fig:4}
\end{figure*}

We run a number of ablations to analyze the influence of different hyperparameters of V-CSS and Q-CSS. Specifically, we conducted all ablations by building on top of ensemble-based model LMH~\cite{clark2019don}. Results are illustrated in Figure~\ref{fig:4}.

\noindent\textbf{The size of $\mathcal{I}$ in V-CSS.} The influence of different size of $\mathcal{I}$ is shown in Figure~\ref{fig:4} (a). We can observe that the model's performance gradually decreases with the increase of $|\mathcal{I}|$. 

\noindent\textbf{The size of critical objects in V-CSS.} The influence of masking different numbers of critical objects is shown in Figure~\ref{fig:4} (a). We compared the dynamic K (Eq.~\eqref{eq:topk_objects}) with some fixed constants (\eg, 1, 3, 5). From the results, we can observe that the dynamic K achieves the best performance.

\noindent\textbf{The size of critical words in Q-CSS.} The influence of replacing different sizes of critical words is shown in Figure~\ref{fig:4} (b). From the results, we can observe that replacing only one word (\ie, top-1) achieves the best performance.

\noindent\textbf{The proportion $\delta$ of V-CSS and Q-CSS.} The influence of different $\delta$ is shown in Figure~\ref{fig:4} (c). From the results, we can observe that the performance is best when $\delta = 0.5$ .

\begin{table}[tbp]
	\small
	\begin{center}
		\scalebox{0.90}{
			\begin{tabular}{l | l | l | c c c c}
				\hline
				& & Model & All & Y/N & Num & Other  \\
				\hline
				\parbox[t]{2mm}{\multirow{5}{*}{\rotatebox[origin=c]{90}{Plain Models}}} & \multirow{5}{*}{UpDn~\cite{anderson2018bottom}} & Baseline & 39.74 & 42.27 & 11.93  & 46.05 \\
				& & Baseline$^\dagger$ & 39.68 & 41.93 & 12.68 & 45.91 \\
				& & ~+Q-CSS & 40.05 & 42.16 & 12.30 & 46.56 \\
				& & ~+V-CSS & 40.98 & 43.12 & 12.28 & 46.86 \\
				& & ~+$\mathcal{CSS}$ & \textbf{41.16} & \textbf{43.96} & \textbf{12.78} & \textbf{47.48} \\
				\hline
				\parbox[t]{2mm}{\multirow{15}{*}{\rotatebox[origin=c]{90}{Ensemble-Based Models}}} & \multirow{5}{*}{PoE~\cite{clark2019don, mahabadi2019simple}} & Baseline & 39.93 & -- & --  & -- \\
				& & Baseline$^\dagger$ & 39.86 & 41.96 & 12.59 & 46.25 \\
				& & ~+Q-CSS & 40.73 & 42.99 & 12.49 & \textbf{47.28} \\
				& & ~+V-CSS & \textbf{49.65} & \textbf{74.98} & \textbf{16.41} & 45.50 \\
				& & ~+$\mathcal{CSS}$& 48.32 & 70.44 & 13.84 & 46.20 \\
				\cline{2-7}
				& \multirow{5}{*}{RUBi~\cite{cadene2019rubi}} & Baseline & 44.23 & -- & --  & -- \\
				& & Baseline$^\dagger$ & 45.23 & 64.85 & 11.83 & 44.11 \\
				& & ~+Q-CSS & 46.31 & \textbf{68.70} & \textbf{12.15} & 43.95 \\
				& & ~+V-CSS & 46.00 & 62.08 & 11.84 & \textbf{46.95} \\
				& & ~+$\mathcal{CSS}$ & \textbf{46.67} & 67.26 & 11.62 & 45.13 \\
				\cline{2-7}
				& \multirow{5}{*}{LMH~\cite{clark2019don}} & Baseline & 52.05 & -- & --  & -- \\
				& & Baseline$^\dagger$ & 52.45 & 69.81 & 44.46 & 45.54 \\
				& & ~+Q-CSS & 56.66 & 80.82 & 45.83 & 46.98 \\
				& & ~+V-CSS & 58.23 & 80.53 & \textbf{52.48} & 48.13 \\
				& & ~+$\mathcal{CSS}$ & \textbf{58.95} & \textbf{84.37} & 49.42 & \textbf{48.21} \\
				\hline
			\end{tabular}
		} 
	\end{center}
	\vspace{-1.5em}
	\caption{Accuracies (\%) on VQA-CP v2 test set of different VQA architectures. $\mathcal{CSS}$ denotes the model with both V-CSS and Q-CSS.$^\dagger$ represents these results are based on our reimplementation.}
	\label{tab:boost_models}
\end{table}

\subsubsection{Architecture Agnostic}

\noindent\textbf{Settings.} Since the proposed CSS is a model-agnostic training scheme, which can be seamlessly incorporated into different VQA architectures. To evaluate the effectiveness of CSS to boost the debiasing performance of different backbones, we incorporated the  CSS into multiple architectures including: \textbf{UpDn}~\cite{anderson2018bottom}, \textbf{PoE} (Product of Experts)~\cite{clark2019don, mahabadi2019simple}, \textbf{RUBi}~\cite{cadene2019rubi}, \textbf{LMH}~\cite{clark2019don}. Especially, PoE, RUBi, LMH are ensemble-based methods. All results are shown in Table~\ref{tab:boost_models}.

\noindent\textbf{Results.} Compared to these baseline models, the CSS can consistently improve the performance for all architectures. The improvement is more significant in the ensemble-based models (\eg, 6.50\% and 9.79\% absolute performance gains in LMH and PoE). Furthermore, when both two types of CSS are used, models often achieve the best performance.

\subsection{Comparisons with State-of-the-Arts}
\subsubsection{Performance on VQA-CP v2 and VQA v2}
\noindent\textbf{Settings.} We incorporated the CSS into model LMH~\cite{clark2019don}, which is dubbed as \textbf{LMH-CSS}, and compared it with the state-of-the-art models on both VQA-CP v2 and VQA v2. According to the backbone of these models, we group them into: 1) \textbf{AReg}~\cite{ramakrishnan2018overcoming}, \textbf{MuRel}~\cite{cadene2019murel}, \textbf{GRL}~\cite{grand2019adversarial}, \textbf{RUBi}~\cite{cadene2019rubi}, \textbf{SCR}~\cite{wu2019self}, \textbf{LMH}~\cite{clark2019don}, \textbf{HINT}~\cite{selvaraju2019taking}. These models utilize the \textbf{UpDn}~\cite{anderson2018bottom} as their backbone. 2) \textbf{HAN}~\cite{malinowski2018learning}, \textbf{GVQA}~\cite{agrawal2018don}, 
\textbf{ReGAT}~\cite{li2019relation}, \textbf{NSM}~\cite{hudson2019learning}. These models utilize other different backbones, \eg, BLOCK~\cite{ben2019block}, BAN~\cite{kim2018bilinear} \etc. Especially, the AReg, GRL, RUBi, LMH are ensemble-based models.

\noindent\textbf{Results.} The results are reported in Table~\ref{tab:SOTA_v2}. When trained and tested on the VQA-CP v2 dataset (\ie, left side of Table~\ref{tab:SOTA_v2}), the LMH-CSS achieves a new state-of-the-art performance over all question categories. Particularly, CSS improves the performance of LMH with a 6.50\% absolution performance gains (58.95\% vs. 52.45\%). When trained and tested on the VQA v2 dataset (\ie, middle side of Table~\ref{tab:SOTA_v2}), the CSS results in a minor drop in the performance by 1.74\% for LMH. For completeness, we further compared the performance drop between the two benchmarks. Different from previous models that suffer severe performance drops (\eg, 23.74\% in UpDn, and 9.19\% in LMH), the LMH-CSS can significantly decrease the performance drop into 0.96\%, which demonstrate that the effectiveness of CSS to further reduce the language biases in VQA.

\subsubsection{Performance on VQA-CP v1}

\noindent\textbf{Settings.} We further compared the LMH-CSS with state-of-the-art models on VQA-CP v1. Similarly, we group these baseline models into: 1) \textbf{GVQA} with SAN~\cite{yang2016stacked} backbone, 2) \textbf{AReg}, \textbf{GRL}, \textbf{RUBi}, and \textbf{LMH} with UpDn backbone.

\noindent\textbf{Results.} Results are reported in Table~\ref{tab:SOTA_v1}. Compared to these baseline models, the LMH-CSS achieves a new state-of-the-art performance on VQA-CP v1. Particularly, the CSS improves the performance of LMH with a 5.68\% absolution performance gains (60.95\% vs. 55.27\%).

\begin{table}[tbp]
	\small
	\begin{center}
		\scalebox{0.98}{
			\begin{tabular}{| l | c c c c|}
				\hline
				Model & All & Yes/No & Num & Other  \\
				\hline
				GVQA~\cite{agrawal2018don}  & 39.23 & 64.72 & 11.87 & 24.86 \\
				UpDn~\cite{anderson2018bottom} & 39.74 & 42.27 & 11.93 & \textbf{46.05} \\
				~~~+AReg$^\dagger$~\cite{ramakrishnan2018overcoming} & 41.17 & 65.49 & 15.48 & 35.48 \\
				~~~+GRL$^\dagger$~\cite{grand2019adversarial} &  45.69 & 77.64 & 13.21 & 26.97 \\
				~~~+RUBi$^\dagger$$^*$~\cite{cadene2019rubi} & 50.90 & 80.83 & 13.84 & 36.02 \\
				~~~+LMH$^\dagger$$^*$~\cite{clark2019don} & 55.27 & 76.47 & 26.66 & 45.68 \\
				\hline
				~~~+\textbf{LMH-CSS} & \textbf{60.95} & \textbf{85.60} & \textbf{40.57} & 44.62  \\
				\hline
			\end{tabular}
		} 
	\end{center}
	\vspace{-1.5em}
	\caption{Accuracies (\%) on VQA-CP v1 test set of state-of-the-art models. $^\dagger$ represents the ensemble-based methods. $^*$ indicates the results from our reimplementation using offical released codes.}
	\label{tab:SOTA_v1}
\end{table}

\begin{table*}
	\small
	\begin{center}
		\scalebox{0.98}{
			\begin{tabular}{| l | l | c | c c c c | c c c c| c c|}
				\hline
				\multirow{2}{*}{Model}  & \multirow{2}{*}{Venue} & \multirow{2}{*}{Expl.} & \multicolumn{4}{c|}{VQA-CP v2 test $\uparrow$} & \multicolumn{4}{c|}{VQA v2 val $\uparrow$} & \multicolumn{2}{c|}{Gap$\Delta$$\downarrow$} \\
				& & & All & Yes/No & Num & Other & All & Yes/No & Num & Other & All & Other \\
				\hline
				HAN~\cite{malinowski2018learning} & \textit{ECCV'18} & & 28.65 & 52.25 & 13.79 & 20.33 & -- & -- & -- & -- & --  & -- \\
				GVQA~\cite{agrawal2018don} & \textit{CVPR'18} & & 31.30 & 57.99 & 13.68 & 22.14 & 48.24 & 72.03 & 31.17 & 34.65 & 16.94 & 12.51 \\
				ReGAT~\cite{li2019relation} & \textit{ICCV'19} & & 40.42 & -- & -- & -- & 67.18 & -- & -- & -- & 26.76 & -- \\
				RUBi~\cite{cadene2019rubi} & \textit{NeurIPS'19} & & 47.11 & 68.65 & 20.28 & 43.18 & 61.16 & -- & -- & -- & 14.05 & -- \\ 
				NSM~\cite{hudson2019learning} & \textit{NeurIPS'19} & & 45.80 & -- & -- & -- & -- & -- & -- & -- & -- & -- \\
				\cline{1-2}
				UpDn~\cite{anderson2018bottom} & \textit{CVPR'18} & & 39.74 & 42.27 & 11.93 & 46.05 & 63.48 & 81.18 & 42.14 & 55.66 & 23.74 & 9.61 \\
				~~~~+AReg$^\dagger$~\cite{ramakrishnan2018overcoming} & \textit{NeurIPS'18} & & 41.17 & 65.49 & 15.48 & 35.48 & 62.75 & 79.84 & 42.35 & 55.16 & 21.58 & 19.68 \\
				~~~~+MuRel~\cite{cadene2019murel} & \textit{CVPR'19} & & 39.54 & 42.85 & 13.17 & 45.04 & -- & -- & -- & --  & -- & -- \\
				~~~~+GRL$^\dagger$~\cite{grand2019adversarial} & \textit{ACL'19} & & 42.33 & 59.74 & 14.78 & 40.76 & 51.92 & -- & -- & -- & 9.59 & -- \\
				~~~~+RUBi$^\dagger$$^*$~\cite{cadene2019rubi} & \textit{NeurIPS'19} & & 45.23 & 64.85 & 11.83 & 44.11 & 50.56 & 49.45 & 41.02 & 53.95 & 5.33 & 9.84 \\
				~~~~+SCR~\cite{wu2019self} & \textit{NeurIPS'19} & & 48.47 & 70.41 & 10.42 & 47.29 & 62.30 & 77.40 & 40.90 & 56.50 & 13.83 & 9.21 \\
				~~~~+LMH$^\dagger$$^*$~\cite{clark2019don} & \textit{EMNLP'19} & & 52.45 & 69.81 & 44.46 & 45.54 & 61.64 & 77.85 & 40.03 & 55.04 & 9.19 & 9.50  \\
				~~~~+\textbf{LMH-CSS} & \textit{CVPR'20} &  & \textbf{58.95} & \textbf{84.37} & \textbf{49.42} & \textbf{48.21} & 59.91 & 73.25 & 39.77 & 55.11 & \textbf{0.96} & \textbf{6.90} \\
				\hline\hline
				~~~~+HINT~\cite{selvaraju2019taking} & \textit{ICCV'19} & HAT & 47.70 & 70.04 & 10.68 & 46.31 & 62.35 & 80.49 & 41.75 & 54.01 & 14.65 & 7.70 \\
				~~~~+SCR~\cite{wu2019self} & \textit{NeurIPS'19} & HAT & 49.17 & 71.55 & 10.72 & 47.49 & 62.20 & 78.90 & 41.40 & 54.30 & 13.03  &  6.81 \\
				~~~~+SCR~\cite{wu2019self} & \textit{NeurIPS'19} & VQA-X & 49.45 & 72.36 & 10.93 & 48.02 & 62.20 & 78.80 & 41.60 & 54.40 & 12.75 & 6.38 \\
				\hline
			\end{tabular}
		} 
	\end{center}
	\vspace{-1.5em}
	\caption[]{Accuracies (\%) on VQA-CP v2 test set and VQA v2 val set of state-of-the-art models. The gap represents the accuracy difference between VQA v2 and VQA-CP v2. $^\dagger$ represents the \emph{ensemble-based} methods. \emph{Expl.} denotes the model has used extra human annotations, \eg, human attention (HAT) or explanations (VQA-X). $^*$ indicates the results from our reimplementation using official released codes.} 
	\label{tab:SOTA_v2}
	\vspace{-1.5em}
\end{table*}

\begin{table*}[tbp]
	\subfloat[Accuracies (\%) on VQA-CP v2 test set.]{
		\tablestyle{2.5pt}{1.05}\begin{tabular}{l|x{22}x{22}x{22}x{22}}
			\hline
				Model & All & Yes/No & Num & Other  \\
			\hline
			SCR  & 48.47 & 70.41 & 10.42 & 47.29 \\
			LMH &52.45 & 69.81 & 44.46 & 45.54  \\
			LMH+SCR & \multicolumn{4}{c}{continued decrease} \\
			LMH+$\mathcal{CSS}$ & 58.95 & 84.37 & 49.42 & 48.21 \\
			\hline\hline
	\end{tabular}}\hspace{3mm}
	\subfloat[$\mathcal{AI}$ score (\%) on VQA-CP v2 test set.]{
		\tablestyle{2.5pt}{1.05}\begin{tabular}{l|x{22}x{22}x{22}x{22}}
			\hline
			Model & Top-1 & Top-2 & Top-3  \\
			\hline
			UpDn & 22.70 & 21.58 & 20.89 \\
			SCR & 27.58 & 26.29 & 25.38 \\
			LMH & 29.67 & 28.06 & 27.04 \\
			LMH+V-CSS & 30.24 & 28.53 & 27.51 \\
			LMH+$\mathcal{CSS}$& \textbf{33.43} & \textbf{31.27} & \textbf{29.86} \\
			\hline\hline
	\end{tabular}}\hspace{3mm}
	\subfloat[\textbf{Left}: $CS(k)$ (\%) on VQA-CP-Rephrasing; \textbf{Right}: $\mathcal{CI}$ score (\%) on VQA-CP v2 test set.]{
		\tablestyle{2.5pt}{1.05}\begin{tabular}{l|x{22}x{22}x{22}x{22}|x{22}}
				\hline
				Model & k=1 & k=2 & k=3 & k=4  & $\mathcal{CI}$ \\
				\hline
				UpDn  & 49.94 & 38.80 & 31.55 & 28.08 & 6.01 \\
				LMH & 51.68 & 39.84 & 33.38 & 29.11 & 7.44 \\
				LMH+Q-CSS & 54.83  & 42.34 & 35.48 & 31.02 & 9.02 \\
				LMH+$\mathcal{CSS}$ & \textbf{55.04} & \textbf{42.78} & \textbf{35.63} & \textbf{31.17} & 
				\textbf{9.03} \\
			\hline\hline
	\end{tabular}}\hspace{3mm}
	\vspace{-1.0em}
	\caption{Quantitative results about the evaluation of the VQA models' visual-explainable and question-sensitive abilities.}
	\vspace{-0.5em}
	\label{tab:vq_ablatives}
\end{table*}

\subsection{Improving Visual-Explainable Ability}

We will validate the effectiveness of CSS to improve the visual-explainable ability by answering the following questions: \textbf{Q1}: Can existing visual-explainable models be incorporated into the ensemble-based framework? \textbf{Q2}: How does CSS improve the model's visual-explainable ability?

\subsubsection{CSS vs. SCR (Q1)}

\noindent\textbf{Settings.} We equipped the existing state-of-the-art visual-explainable model SCR~\cite{wu2019self} into the LMH framework, and compared it with CSS. Results are reported in Table~\ref{tab:vq_ablatives} (a).

\noindent\textbf{Results.} Since the training of all SOTA visual-explainable models (\eg, SCR, HINT) are not end-to-end, for fair comparisons, we used a well-trained LMH (\ie, 52.45\% accuracies on VQA-CP v2) as the initial model. However, we observe that its performance continues to decrease from the start, which shows that the existing visual-explainable models can not be easily incorporated into the ensemble-based framework. In contrast, CSS can improve the performance.

\subsubsection{Evaluations of Visual-Explainable Ability (Q2)}

\begin{figure*}[htbp]
	\centering
	\includegraphics[width=0.95\linewidth]{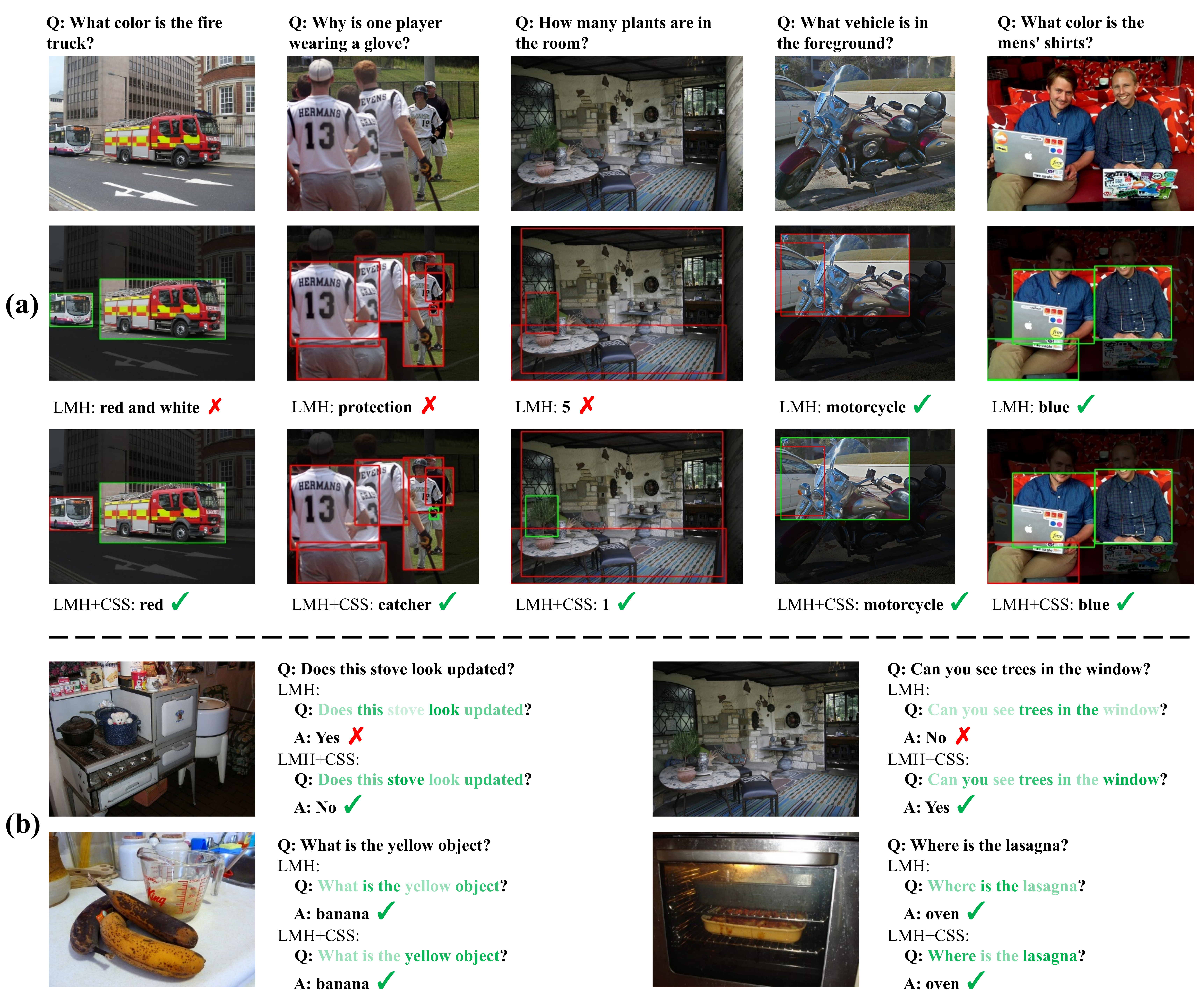}
	\vspace{-0.5em}
	\caption{(a) \textbf{visual-explainable ability}: The \textcolor{green}{\textbf{green}} boxes denote their scores $s(\hat{a}, \bm{v}) \textgreater 0$, \ie, positive contributions to final predictions; The \textcolor{red}{\textbf{red}} boxes denote their scores $s(\hat{a}, \bm{v}) \textless 0$, \ie, negative contributions to final predictions. Only objects which are highly related to the QA pair are shown (\ie, $\mathcal{SIM} \geq 0.6$). (b) \textbf{question-sensitive ability}: The different shades of green color in the question denotes the relative values of $s(\hat{a}, \bm{w})$. Thus, the word with darker green denotes the word has larger contribution to final predictions.}
	\label{fig:5}
\end{figure*}

\noindent\textbf{Settings.} We evaluate the effectiveness of CSS to improve the visual-explainable ability on both quantitative and qualitative results. For quantitative results, since we lack human annotations about the critical objects for each question, we regard the $\mathcal{SIM}$ score (Section~\ref{sec:v-css} \textsc{IO\_Sel}) as pseudo ground truth. Thus, we design a new metric \emph{Average Importance} ($\mathcal{AI}$): the average $\mathcal{SIM}$ score of the top-K objects with highest $|s(a, \bm{v})|$. The results are shown in Table~\ref{tab:vq_ablatives} (b). For qualitative results, we illustrated in Figure~\ref{fig:5} (a).

\noindent\textbf{Results.} From Table~\ref{tab:vq_ablatives} (b), we can observe that CSS dramatically improves the $\mathcal{AI}$ scores, which means the actually influential objects are more related to the QA pair. From Figure~\ref{fig:5} (a), we can find that the CSS helps the model to make predictions based on critical objects (\ie, green boxes), and suppress the influence of irrelevant objects (\ie, red boxes).

\subsection{Improving Question-Sensitive Ability}
We will validate the effectiveness of CSS to improve the question-sensitive ability by answering the following questions: \textbf{Q3}: Does CSS helps to improve the robustness to diverse rephrasings of questions? \textbf{Q4}: How does CSS improve the model's question-sensitive abilities?

\subsubsection{Robustness to Rephrasings of Questions (Q3)}
\noindent\textbf{Settings.} As discussed in previous work~\cite{shah2019cycle}, being robust to diverse rephrasing of questions is one of key behaviors of a question-sensitive model. To more accurately evaluate the robustness, we re-splited the existing dataset VQA-Rephrasings~\cite{shah2019cycle} with the same splits as VQA-CP, and denoted it as VQA-CP-Rephrasings. For evaluation, we used the standard metric \emph{Consensus Score} $CS(k)$. Results are reported in Table~\ref{tab:vq_ablatives} (c) (left). We refer readers to~\cite{shah2019cycle} for more details about the VQA-Rephrasings and metric $CS(k)$.

\noindent\textbf{Results.} From Table~\ref{tab:vq_ablatives} (c), we can observe that Q-CSS dramatically improves the robustness to diverse rephrasings of questions. Furthermore, V-CSS can help to further improve the robustness, \ie, $\mathcal{CSS}$ achieves the best performance.

\subsubsection{Evaluations of Question-Sensitive Ability (Q4)}

\noindent\textbf{Settings.} We evaluate the effectiveness of CSS to improve the question-sensitive ability on both quantitative and qualitative results. For quantitative results, since there is no standard evaluation metric, we design a new metric \emph{Confidence Improvement} ($\mathcal{CI}$): Given a test sample $(I, Q, a)$, we remove a critical noun in question $Q$, and obtain a new test sample $(I, Q^*, a)$\footnote{The auxiliary test set is released in: \href{https://github.com/yanxinzju/CSS-VQA}{github.com/yanxinzju/CSS-VQA}}. Then we feed both two samples into evaluated model, and calcluate the confidence decreses of the ground-truth answer. We formally define $\mathcal{CI}$ in Eq.~\ref{eq:CI}:
\begin{equation} \label{eq:CI}
\small
\mathcal{CI} = \frac{\sum_{(I, Q)}  (P_{vqa}(a | I, Q) - P_{vqa}(a | I, Q^*)) \cdot \mathbf{1}(a = \hat{a}) }{\sum_{(I, Q)} 1}
\end{equation}
where $\hat{a}$ is the model predicted answer for sample $(I, Q)$, $\mathbf{1}$ is an indicator function. The results are reported in Table~\ref{tab:vq_ablatives} (c). For qualitative results, we illustrated in Figure~\ref{fig:5} (b).

\noindent\textbf{Results.} From Table~\ref{tab:vq_ablatives} (c), we can observe that CSS helps the model to benefit more from the critical words, \ie, removing critical words results in more confidence drops for the ground-truth answers. From Figure~\ref{fig:5} (b), we can find that CSS helps the model to make predictions based on critical words (\eg, ``stove" or ``lasagna"), \ie, forcing model to understand the whole questions before making predictions.

\section{Conclusion}
In this paper, we proposed a model-agnostic Counterfactual Samples Synthesizing (CSS) training scheme to improve the model's visual-explainable and question-sensitive abilities. The CSS generates counterfactual training samples by masking critical objects or words. Meanwhile, the CSS can consistently boost the performance of different VQA models. We validate the effectiveness of CSS through extensive comparative and ablative experiments. Moving forward, we are going to 1) extend CSS to other visual-language tasks that suffer severe language biases; 2) design a specific VQA backbone to benefits from CSS.

\footnotesize \noindent\textbf{Acknowledgement} This work was supported by National Key Research \& Development Project of China (No.2018AAA0101900), National Natural Science Foundation of China (U19B2043, 61976185), Zhejiang Natural Science Foundation (LR19F020002, LZ17F020001), Fundamental Research Funds for the Central Universities and Chinese Knowledge Center for Engineering Sciences and Technology. Long Chen was supported by 2018 ZJU Academic Award for Outstanding Doctoral Candidates.

{\small
\bibliographystyle{ieee_fullname}
\bibliography{cvpr20}
}

\end{document}